%% file: naaclhlt2018.tex
\documentclass[11pt,a4paper]{article}
\usepackage[table]{xcolor}
\usepackage[hyperref]{naaclhlt2018}
\usepackage{times}
\usepackage{latexsym}
\usepackage{graphicx}
\usepackage{booktabs}
\usepackage{url}
\usepackage{comment}
\usepackage{multirow}
\usepackage{array}
\usepackage{graphicx}
\usepackage{multirow}

\usepackage[english]{babel}
\usepackage[autostyle, english=american]{csquotes}
\MakeOuterQuote{"}

\aclfinalcopy 
\def\aclpaperid{1063} 

\newcommand\MyBox[2]{
  \fbox{\lower0.75cm
    \vbox to 1.7cm{\vfil
      \hbox to 1.7cm{\hfil\parbox{1.4cm}{#1\\#2}\hfil}
      \vfil}%
  }%
}

\title{Detecting Linguistic Characteristics of Alzheimer's Dementia \\ by Interpreting Neural Models}

\author{Sweta Karlekar $\;\;\;\;\;\;\;\;$ Tong Niu $\;\;\;\;\;\;\;\;$ Mohit Bansal \\
  UNC Chapel Hill \\
  {\tt \{swetakar, tongn, mbansal\}@cs.unc.edu}   }

\date{}

\begin{document}
\maketitle

\input{tex-files/abstract}
\input{tex-files/introduction}
\input{tex-files/related-works}
\input{tex-files/models}

\input{tex-files/exp-setup}
\input{tex-files/results}
\input{tex-files/analysis}
\input{tex-files/conclusion}

\vspace{-5pt}
\section*{Acknowledgments}
\vspace{-5pt}
We thank the anonymous reviewers for their helpful comments. This work was supported by a
Google Faculty Research Award, a Bloomberg Data Science Research Grant, an IBM Faculty
Award, and NVidia GPU awards.

\bibliography{references}
\bibliographystyle{acl_natbib}

\input{tex-files/supplementary}

\end{document}

%% file: tex-files/abstract.tex
\begin{abstract}
Alzheimer's disease (AD) is an irreversible and progressive brain disease that can be stopped or slowed down with medical treatment. 
Language changes serve as a sign that a patient's cognitive functions have been impacted, 
potentially leading to early diagnosis. 
In this work, we use NLP techniques to classify and analyze the linguistic characteristics of AD patients using the DementiaBank dataset. 
We apply three neural models based on CNNs, LSTM-RNNs, and their combination, 
to distinguish between language samples from AD and control patients. 
We achieve a new independent benchmark accuracy for the AD classification task. 
More importantly, we next interpret what these neural models have learned about the linguistic characteristics of AD patients, 
via analysis based on activation clustering and first-derivative saliency techniques. 
We then perform novel automatic pattern discovery inside activation clusters, and consolidate AD patients' distinctive grammar patterns. 
Additionally, we show that first derivative saliency can not only rediscover previous language patterns of AD patients, 
but also shed light on the limitations of neural models.
Lastly, we also include analysis of gender-separated AD data.
\end{abstract}

%% file: tex-files/introduction.tex
\section{Introduction}
Alzheimer's dementia is the most common form of dementia, caused by Alzheimer's disease (AD). 
AD cannot be cured or reversed~\cite{glenner1990alzheimer}. 
However, medication can be used to slow or halt degeneration especially when detected at an early stage. Current diagnoses often involve lengthy medical evaluations. 
One of the early symptoms of AD, cognitive impairment---which can be evidenced by issues with word-finding, impaired reasoning or judgment, and changes in language~\cite{mckhann1984clinical}---is
motivating linguists and computer scientists 
to help quickly diagnose people afflicted by this disease.

This task is challenging because it requires diverse linguistic and world knowledge. 
For example, the sentence "Well...there's a mother standing there uh uh washing the dishes and the sink is overspilling...overflowing" is AD-positive. 
To distinguish this from a control sample, 
one needs to know that the word "overspill" is not common in American English~\cite{davies2009385+}, 
and the speaker is correcting themselves by saying "overspilling...overflowing", which hints on signs of confusion and memory loss~\cite{duke2002cognitive}. 
Moreover, different grammar patterns emerge based on the scenario at hand.
In addition, the characteristics of AD-affected speech vary between stages of disease progression~\cite{konig2015automatic}, making it harder for feature-based approaches to adapt.

Motivated by the shortcomings of manual feature-engineering for such a diverse and complex task, 
we first present three end-to-end neural models to address it. 
The first two are the widely adopted CNN and LSTM-RNN models, 
and the third is a stronger joint CNN-LSTM architecture. 
Our best-performing model requires only minimal feature engineering (namely automatic, 
commonly-used POS tags that are already present in the dataset) and establishes a new independent benchmark that outperforms previous AD classification scores. 

More importantly, we next present interpretation results to explain what AD-relevant linguistic features these neural models are learning, 
via two visualization techniques: Activation Clustering~\cite{girshick2014rich} and First Derivative Saliency~\cite{simonyan2013deep}, 
plus our novel approach of automatically discovering grammatical patterns common in different activation clusters. 
Furthermore, we split our dataset by gender and analyze the performance of our model on each subsample of the data to illustrate that the features we find are not gender-specific.
These methods not only help rediscover AD linguistic features that have been found by previous works---including short answers, bursts of speech, repeated requests for clarification, and starting with interjections---but also lead to new insights in AD characteristics via our automatic speech pattern extraction method.
These findings could potentially help improve the accuracy and speed of medical diagnoses. 

\begin{figure}[t]
\centering
\includegraphics[width=0.5\textwidth]{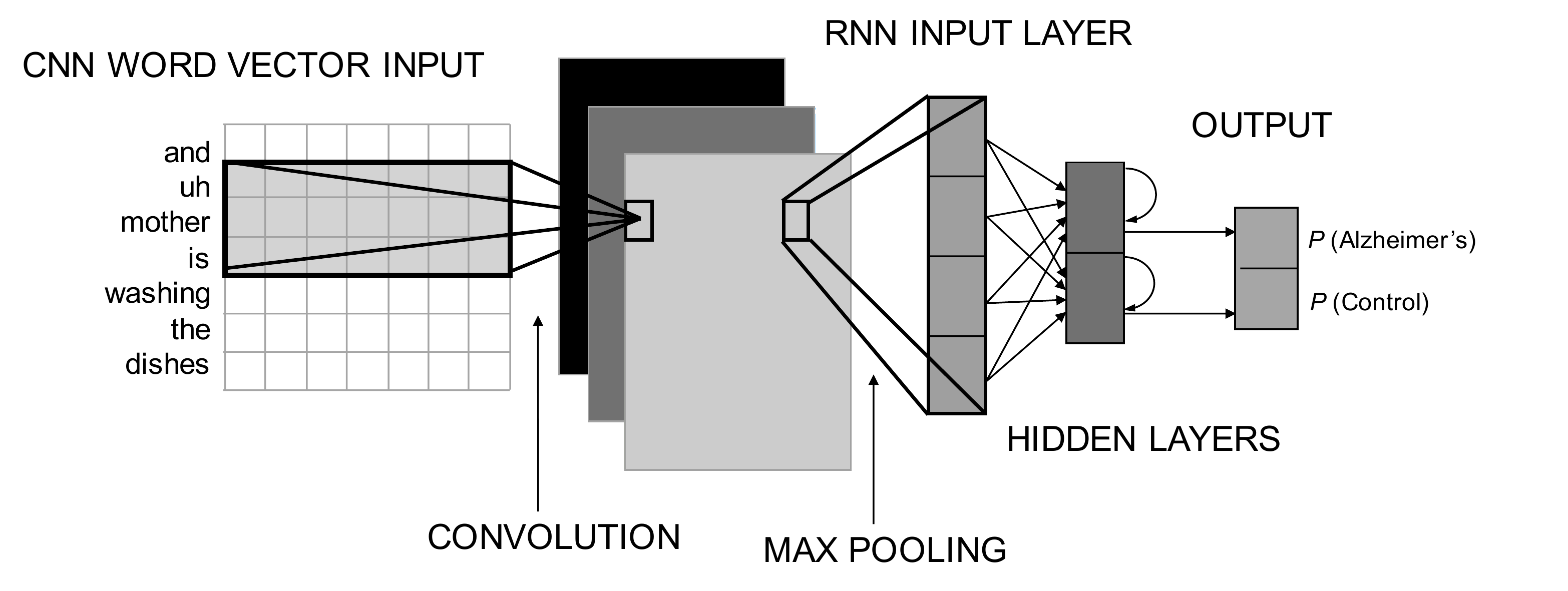}
\vspace{-28pt}
\caption{\label{fig:cnnrnn} Our CNN-LSTM hybrid neural network.}
\label{fig: CNN-LSTM}
\end{figure}

%% file: tex-files/related-works.tex
\section{Related Works}
\label{sect: Related Works}

\noindent\textbf{Language-based Alzheimer's Detection}:
\label{subsect: Approaches on Alzheimer Language Detection}
Previous works using language to detect AD relied mainly 
on hand-crafted features from transcripts ~\cite{orimaye2017predicting,orimaye2015learning}, 
occasionally using acoustic data~\cite{konig2015automatic,rudzicz2014automatically}.
The challenge with feature-based approaches is that they rely heavily on the researchers' linguistics and medical expertise, 
and are also hard to generalize to other progression stages and disease types, 
which may correspond to different linguistic features. Hand-picked features may also become outdated as language and culture evolves. 
Moreover, some features may be too nuanced for humans to detect, especially at early stages of AD. 
In order to address these issues,~\newcite{orimaye2016deep} adopted a deep neural network language model. 
However, a neural approach is usually a black-box and it is hard to interpret its reasoning for the final classification decisions. 
To make our approaches more interpretable while harvesting the benefits of neural approaches,
we present three accurate neural models and include multiple visualization techniques to illustrate both their effectiveness and limitations. 

\noindent\textbf{CNN-LSTM on NL Classification}:
\label{subsect: LSTM and CNN on Natural Language Classification}
CNN and LSTM~\cite{hochreiter1997long} have both been leveraged extensively for extracting features in natural language.
Our best performing model CNN-LSTM is closely related to C-LSTM by~\newcite{zhou2015c}, where an LSTM is laid on top of a CNN model. 
This model has been shown to perform better at sentiment classification than either of its integral parts alone.

\noindent\textbf{Visualization Techniques for Neural Models}:
There have been various visualization techniques proposed for neural networks in both Computer Vision~\cite{krizhevsky2012imagenet,simonyan2013deep,zeiler2014visualizing,samek2017evaluating,mahendran2015understanding} and NLP~\cite{li2015visualizing,kadar2017representation}. In this work, we adopt two visualization techniques: Activation Clustering~\cite{girshick2014rich} following the politeness interpretation work of~\newcite{aubakirova2016interpreting}, which leads to insight on sentence-level patterns, and First Derivative Saliency~\cite{simonyan2013deep} following~\newcite{li2015visualizing} and ~\newcite{aubakirova2016interpreting}, which 
provides insight to the importance of each word in deciding the final classification label. 

%% file: tex-files/models.tex
\section{Classification Models}
\label{sect: Classification Models}

\noindent\textbf{CNN}:
For each sentence, we apply an embedding and a convolutional layer, followed by a max-pooling layer~\cite{collobert2011natural}. The convolution features are obtained by applying filters of varying window sizes to each window of words. The result is then passed to a softmax layer that outputs probabilities over two classes.

\noindent\textbf{LSTM-RNN}:
CNNs are not specialized for capturing long-range sequential correlations~\cite{pascanu2013construct}.
We thus also experimented with an LSTM-RNN model, 
which consists of an embedding layer followed by an LSTM layer.
The final state, containing information from the entire sentence, 
is fed to a fully-connected layer followed by a softmax layer to obtain the output probabilities.

\noindent\textbf{CNN-LSTM}:
Observing that both models achieve results comparable to previous best performing approach, 
and considering that they each have their own complementary strengths, we experimented with a combined architecture, laying an LSTM layer on top of CNN (See Figure~\ref{fig: CNN-LSTM}). This CNN layer is identical to the vanilla CNN before the max-pooling layer, and the LSTM layer is identical to the vanilla LSTM-RNN after the embedding layer. More details are provided in the appendix.

%% file: tex-files/exp-setup.tex
\section{Experimental Setup}
\label{sect: Experimental Setup}

\begin{figure}[t]
\centering
\small
\includegraphics[width=0.4\textwidth]{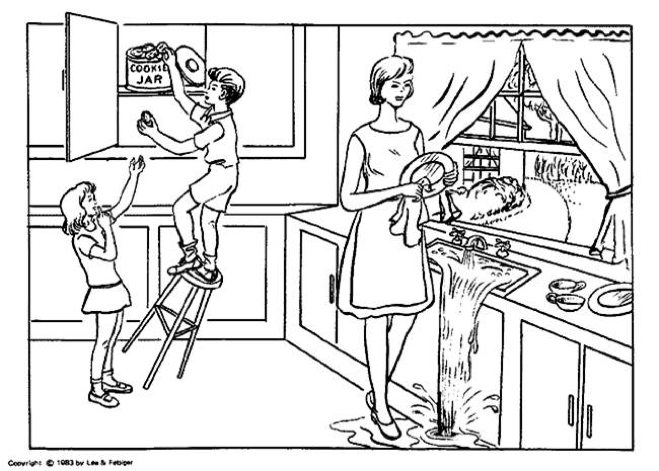}
\vspace{-10pt}
\caption{\label{fig:cookie} Boston cookie theft description task. Participants were asked to describe all events in the image.}
\end{figure} 

\paragraph{Dataset}
\label{para:Dataset}
This study utilizes DementiaBank~\cite{boller2005dementiabank}, the largest publicly available dataset of transcripts and audio recordings of AD (and control) patient interviews.\footnote{\url{http://dementia.talkbank.org}} Patients were asked to perform various tasks; for example, in the "Boston Cookie Theft" description task, patients were shown an image and asked to describe what they see (See Figure~\ref{fig:cookie}). 
Other tasks include the `Recall Test' in which patients were asked to recall attributes of a story they had been told previously.
Each transcript in DementiaBank comes with automatic
morphosyntactic analysis, 
such as standard part-of-speech tagging, 
description of tense, and repetition markers.\footnote{\url{http://talkbank.org/manuals/CHAT.docx}} Note that these features are generic, automatically-extracted linguistic properties and are not AD-specific.
We broke each transcript into individual utterances to use as data samples. 
Note that we also removed utterances that did not have accompanying POS tags. 
This balancing reduced the amount of data but ensured fair comparison between models with tagged and untagged setups.

\paragraph{Training Details}
Our CNN model was a $2$-D convolutional neural network. Filter sizes of [$3$, $4$, $5$] were used. 
Our LSTM-RNN had $2$ layers. The CNN-LSTM model had filter sizes [$3$, $4$, $5$, $6$] and $1$ LSTM hidden layer. 
For each model, all hyperparameters were tuned using the dev set. 

See appendix for dataset and training details. 

%% file: tex-files/results.tex
\section{Results}
\label{sect: Results}
With untagged data, our CNN, LSTM and CNN-LSTM models achieved an accuracy of $82.8\%$, $83.7\%$ and $84.9\%$, respectively. 
When fed with the given POS-tagged data, our best-performing CNN-LSTM model achieved $91.1\%$ in accuracy, 
setting a new benchmark accuracy for this task (see Table~\ref{tab:widgets} for more details).\footnote{We also tried using a bidirectional RNN, which gave 84.7\%, 86.2\% and 91.1\% accuracies for our LSTM-RNN, CNN-RNN, and CNN-RNN-tagged models, respectively.}
Compared to other related works, ~\newcite{orimaye2015learning,orimaye2017predicting} used AUC instead of accuracy, 
and~\newcite{konig2015automatic} did not use DementiaBank data.~\newcite{rudzicz2014automatically} achieved an accuracy of 67.0\% on the DementiaBank dataset using audio features as well as transcripts. 
\newcite{orimaye2016deep} achieved an accuracy of $87.5\%$ but only used $36$ transcripts. 
Their test set is thus different from ours, 
leading to very little data and high variance. 
To the best of our knowledge,~\newcite{orimaye2016deep} used the original transcripts, which include POS tags. 
Hence, most previous works on this topic are not directly comparable to our work, but we aim to establish a new independent, strong neural benchmark and then more importantly focus on visualization and interpretability of neural models.

\begin{table}[t]
\centering
\begin{small}
\begin{tabular}{r|r|l}
Model & Details & Accuracy \\\hline
2D-CNN & Non-Tagged Utterances & 82.8 \\
LSTM & Non-Tagged Utterances & 83.7 \\ 
CNN-RNN & Non-Tagged Utterances & 84.9 \\
CNN-RNN & POS-tagged Utterances & \textbf{91.1}
\end{tabular}
\end{small}
\caption{\label{tab:widgets} Accuracy results of models. Note that we downsampled the data to remove utterances that did not have accompanying POS tags, so as to allow fair comparison between the tagged and untagged models.
}
\end{table}

Based on error analysis of the POS-based CNN/RNN model's classification result, 
we found that almost all AD-positive results are classified correctly as AD-positive. 
However, there is more error in classifying non-AD samples, 
which could be due to the fact that DementiaBank includes patients with probable and possible AD, 
each exhibiting various degrees of symptoms. 
Patients who are AD-positive may still have partially unaffected speech (similar to non-AD control patients' speech). However, because all utterances from AD-positive interviews are tagged as AD-positive, 
these seemingly unaffected utterances are still tagged AD-positive. 
To further understand the errors of our model, 
$10\%$ of the wrongly classified non-AD examples were randomly selected and analyzed.
Of this smaller sample, $36.3\%$ were short utterances such as \textit{"okay"}, \textit{"alright"}, \textit{"oh my"}, etc. 
These forms of speech are utterances that are present in both classes, 
but more commonly found in AD-positive cases. 
The remaining $63.7\%$ were examples of speech that could be classified either way without surrounding context, 
such as "\textit{“she's drying dishes”}" and "\textit{“stool's tipping over”}". 
Hence, future work could incorporate context from surrounding samples in each interview to help distinguish between temporarily unaffected speech patterns in AD-positive patients and the continually unaffected speech of non-AD control patients.

%% file: tex-files/analysis.tex
\begin{table}[t]
  \centering
  \small
    \begin{tabular}{rl|rl|rl}
    \multicolumn{6}{c}{Non-AD Clusters - Cookie Task} \\ \hline
    POS   & Freq & \multicolumn{1}{l}{POS} & Freq & POS   & Freq \\ \hline
    \textit{n} & 0.15  & \textit{n} & 0.13 & \textit{n} & 0.15\\
    \textit{det} & 0.13  & \textit{det} & 0.13 & \textit{det} & 0.13\\
    \textit{presp} & 0.07  & \textit{part} & 0.09 & \textit{part} & 0.10\\
    \textit{part} & 0.05  & \textit{presp} & 0.09 & \textit{presp} & 0.10\\
    \end{tabular}%
    \vspace{-5pt}
    \caption{ \label{tab:nonadclusters} Top POS tags and frequencies for three non-AD clusters for the Cookie task.}
    \vspace{5pt}
\end{table}%

\begin{figure*}[!htb]
\minipage{0.33\textwidth}
  \includegraphics[width=\linewidth]{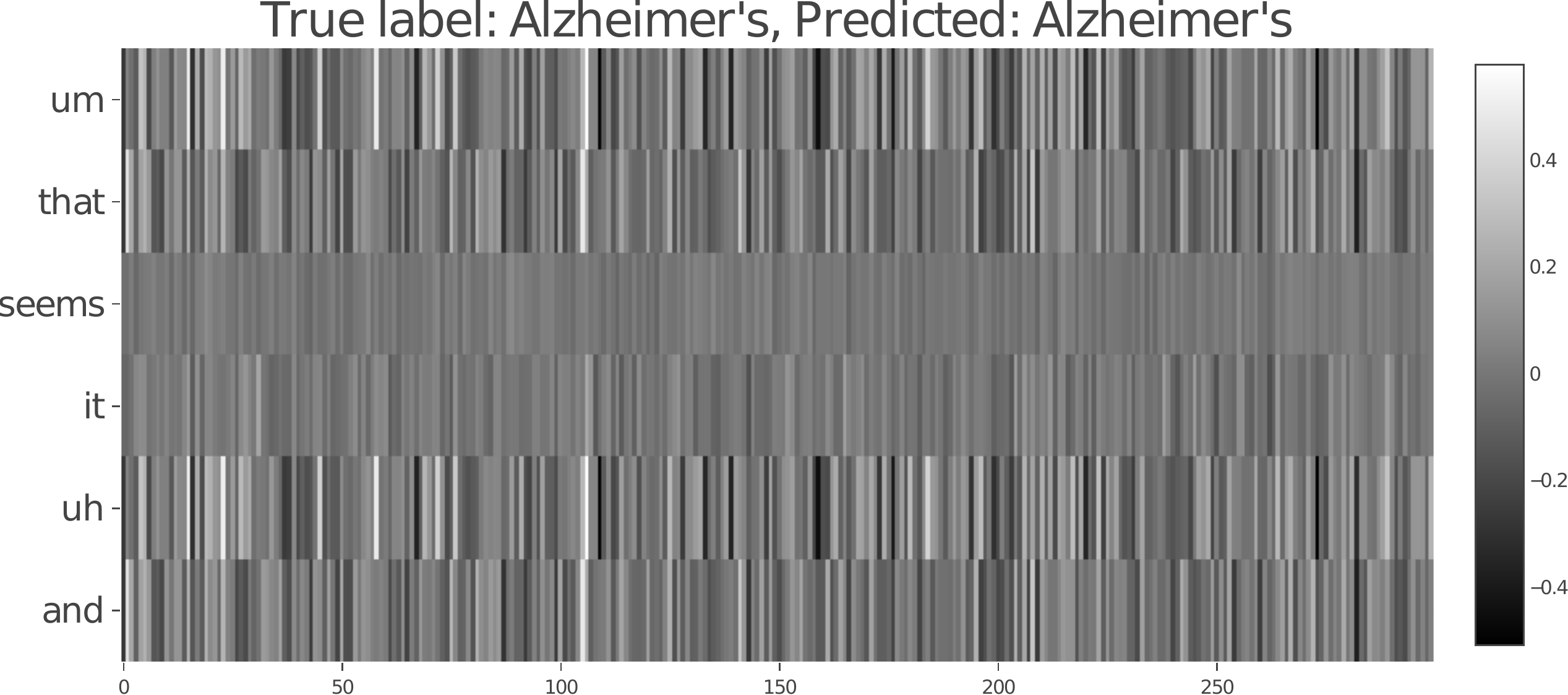}
\endminipage\hfill
\minipage{0.33\textwidth}
  \includegraphics[width=\linewidth]{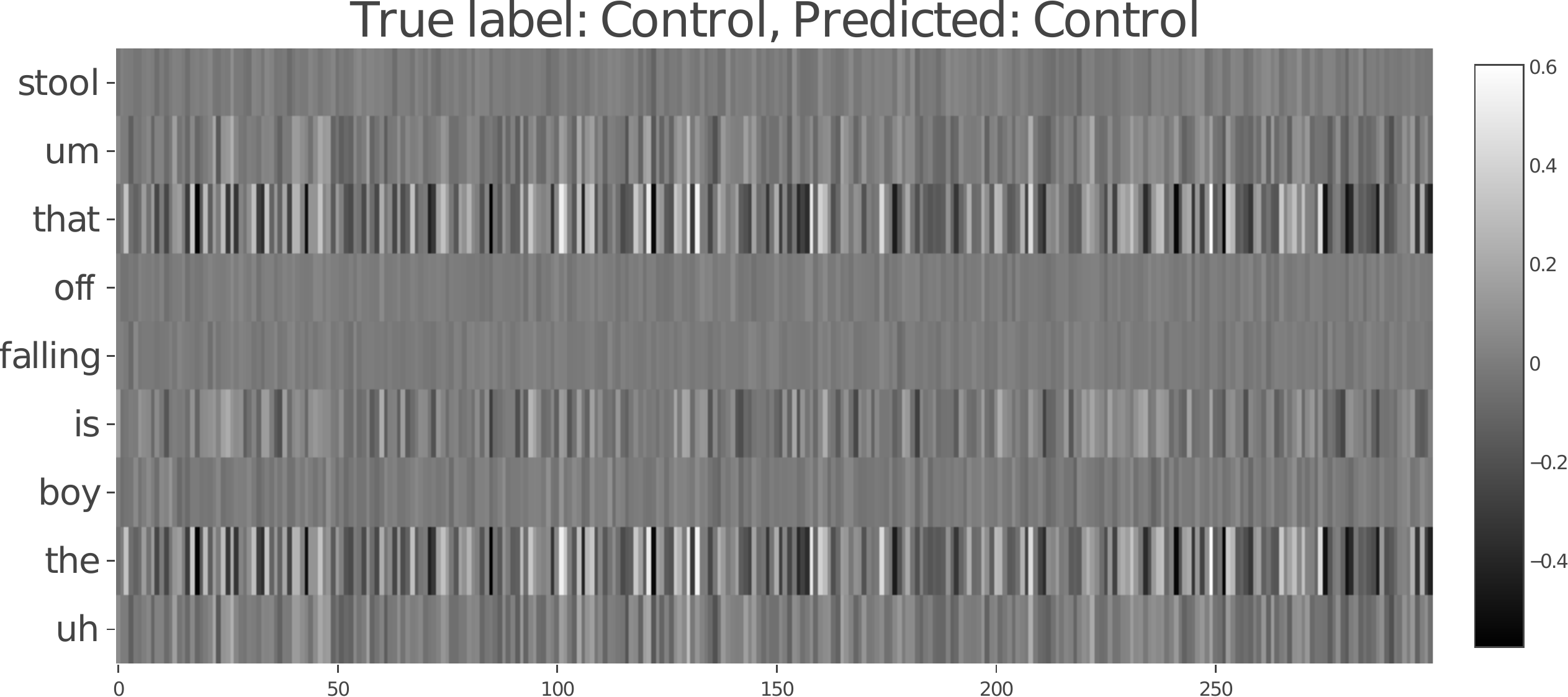}
\endminipage\hfill
\minipage{0.33\textwidth}%
  \includegraphics[width=\linewidth]{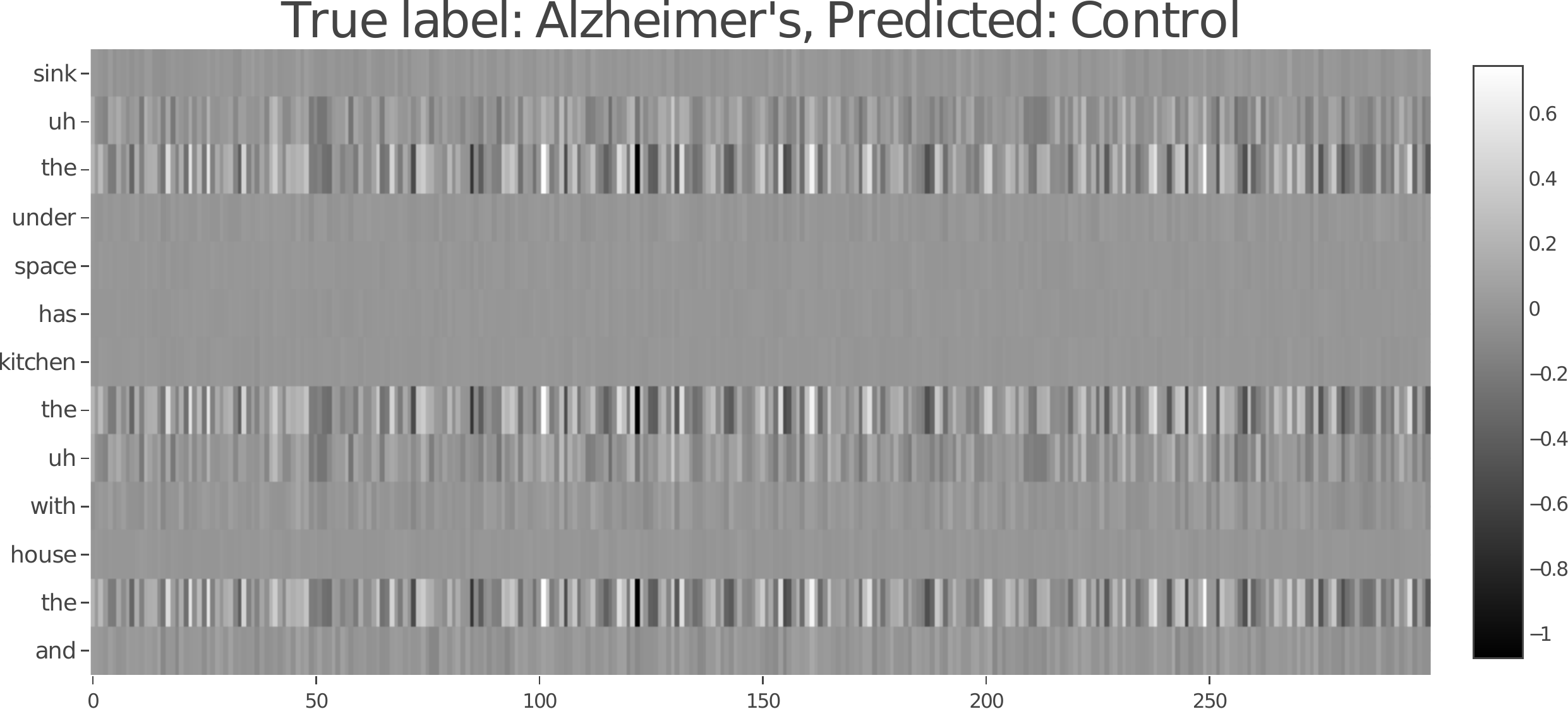}
\endminipage
\vspace{-8pt}
\caption{
\textbf{Left + Middle}:
first derivative saliency heat maps for correctly classified Alzheimer's and control examples. 
\textbf{Right}: first derivative heat saliency maps for incorrectly classified Alzheimer's example. 
}
\label{fig:examples}
\vspace{-10pt}
\end{figure*}

\section{Analysis and Visualization}
We first present gender analysis of our results and then present interpretation of the linguistic cues our CNN-LSTM model identified via two visualization strategies: activation clustering~\cite{girshick2014rich} and first-derivative saliency heat maps~\cite{simonyan2013deep}.

\subsection{Gender Differences}
Many previous works have debated on the difference in language for male versus female patients with Alzheimer's \cite{bayles1999gender, mcpherson1999gender, buckwalter1996gender, ripich1995gender, hebert2000decline,heun2002gender}. 
In agreement with some of these previous works, we found that the sets of the top ten most common POS-tags for both AD-positive men and women are the same, i.e., we did not detect a significant difference in the language complexity or syntax of male and female patients with AD in our dataset. 
Moreover, our best performing model achieved 86.6\% classification accuracy on solely the male data and 86.2\% accuracy on solely the female data, demonstrating that it found no statistically significant difference between the AD-positive language of men versus women.\footnote{Statistical insignificance calculated using the bootstrap test (Noreen, 1989; Efron and Tibshirani, 1994). 
We split our dataset based on gender and down-sampled the female data subset such that it had the same data-size and non-AD to AD data-ratio as the male data subset.}

\subsection{Activation Clusters}
Activation clustering~\cite{girshick2014rich} treats the activation values of \textit{n} neurons per input as coordinates in an \textit{n}-dimensional space. 
K-means clustering is then performed to group together inputs that maximally activate similar neurons.

\subsubsection{Rediscovering Existing Strategies}
\label{subsubsect: Rediscovering Existing Strategies}
Our activation clusters corroborated previous studies, forming clusters around known linguistic characteristics of Alzheimer's disease \cite{watson1999analysis,rudzicz2014automatically}. 

\noindent{\bf{Short Answers and Bursts of Speech}}
Clusters formed around short answers, which have been split up by natural pauses in speech.
\textit{{\{`okay', `and', `yes', `oh !', `yes', `fine'\}}}

\noindent{\bf{Repeated Requests for Clarification}}
Another cluster formed around clarification questions and confusion about the task, specifically in the past tense.
\textit{{\{`Did I say facts ?', `Did I get any ?', `Did I say elephant ?'\}}}

\noindent{\bf{Starting with Interjections}}
Many clusters contain utterances that start with interjections such as "oh", "well", "so", and "right". 
\textit{{\{`Well I gotta see it', `Oh I just see a lot of uh...', `So all the words that you can...'\}}}

\begin{table}[t]
\small
  \centering
    \begin{tabular}{rl|rl}
    \multicolumn{2}{c|}{AD} & \multicolumn{2}{c}{Non-AD} \\ \hline
    POS   & Frequency & POS   & Frequency \\ \hline
    \textit{n} & 0.20  & \textit{n} & 0.15 \\
    \textit{det} & 0.14  & \textit{det} & 0.13 \\
    \textit{adj} & 0.05  & \textit{presp} & 0.07 \\
    \textit{adv} & 0.04  & \textit{part} & 0.05 \\
    \end{tabular}%
    \vspace{-5pt}
\caption{\label{tab:postags} Top POS tags in AD cluster and non-AD cluster for Cookie task.}
\vspace{5pt}
\end{table}%

\subsubsection{Automatic Cluster Pattern Analysis}
Next, we extend activation clustering to perform novel automatic pattern discovery inside different clusters, as opposed to manually looking for patterns as in~\newcite{aubakirova2016interpreting}.
Finding the most common POS tags in each cluster allows us to better understand which grammatical structures are favored. 
No two clusters had exactly the same most-common POS tags, but many clusters shared similar top POS-tags on the same task.

\paragraph{Control Clusters}
An example unaffected speech cluster for the Cookie task has the following as the most common POS tags: [\textit{(`n', $.15$), (`det', $.13$), (`presp', $.07$), (`part', $.05$)}].\footnote{POS tags used in this paper: v=verb, n=noun, pro=pronoun, adv=adverb, det=determiner, aux=auxiliary verb, prep=preposition, co=interjection, part=participle, presp=present participle. 
The frequencies of each POS tag are scaled based on the total number of tags in each cluster.} 
This pattern follows other clusters of unaffected speech, with nouns, determiners, and participles always found in the most-used POS tags. 
To illustrate this, two other control clusters found for the Cookie task have very similar top POS tags and frequencies: [\textit{(`n', $.13$), (`det', $.13$), (`part', $.09$), (`presp', $.09$)}] and [\textit{(`n', $.15$), (`det', $.13$), (`part', $.10$), (`presp', $.10$)}] (see Table \ref{tab:nonadclusters} for side-by-side comparison of these clusters). 
For the same task, non-AD clusters contain the same top four POS tags.

\paragraph{AD Clusters}
For the Recall task, one cluster of AD patients' speech shows that the most common POS tags are [\textit{(`n', $.15$), (`co', $.15$), (`v', $.06$), (`pro', $.06$)}].
Across samples from the Recall task, AD-marked clusters contained frequent interjections and verbs.
On the other hand, in regards to the Cookie task, the most common POS tags for the AD cluster found are [\textit{(`n', $.20$), (`det', $.14$), (`adj', $.05$), (`adv', $.04$)}], i.e., more adjectives and adverbs. Hence, between different tasks such as Cookie and Recall, the most commonly used POS tags for AD clusters are distinct. 

Moreover, qualitative analysis shows dissimilarities in the most common POS tags between the Cookie task's AD and non-AD cluster(s). Table \ref{tab:postags} shows the comparison between representative AD and non-AD clusters for the Cookie task. The AD-positive cluster has only 2 most-used POS tags in common with the non-AD cluster. In fact, none of the 3 non-AD clusters found in Table \ref{tab:nonadclusters} have adjectives or adverbs in their most-used POS tags list, unlike the AD cluster in Table \ref{tab:postags}.

\subsection{First Derivative Saliency Heat Maps}
Saliency heat maps~\cite{simonyan2013deep} illustrate which words in an input had the biggest impact on the classification of the whole sentence. 
This is done by taking the gradient of the final scores w.r.t. the word embeddings of the inputs. 

\paragraph{Heat Map Analysis}
The filler words "uh" and "um" are emphasized in Figure~\ref{fig:examples} (left), showing that they have a lot of influence on classification. 
The initial "and" is highlighted as well, corroborating the results of the activation clusters in that starting with a coordinating conjunction is a trait of Alzheimer's speech. 
However, in Figure~\ref{fig:examples} (middle), the "uh" filler word is not highlighted, showing that most control patients do not use filler words as heavily as Alzheimer's patients. Instead, words that give structure to a sentence have the biggest impact on classification, such as definite articles and determiners (e.g., "the" and "that"). 
Figure~\ref{fig:examples} (middle) shows that the most highlighted words are "the", "that" and "is". 

\paragraph{Visualizing Limitations}
Furthermore, by visualizing an incorrectly classified example, we can learn about the limitations of our neural network. 
Figure~\ref{fig:examples} (right) illustrates a map of an incorrectly predicted sample.
The model misclassified it due to the length of the utterance (activation clustering showed that AD patients tend to have short bursts of speech, see Section~\ref{subsubsect: Rediscovering Existing Strategies}) and the heavy use of determiners. 
However, the repeated "um"s and starting with a coordinating conjunction strongly indicated AD, which confused our model. 
From Figure~\ref{fig:examples}, 
the repeated use of filler words (i.e. "uh") had the second most influence on classification. 
In future work, with more context data and advanced neural methods, our model's next steps will be to learn how to better classify samples that strongly exhibit both AD and control features. 

%% file: tex-files/conclusion.tex
\vspace{-5pt}
\section{Conclusion}
\vspace{-5pt}
\label{sect: Conclusion}
We applied three models to the AD classification task, and our CNN-LSTM model achieves a new benchmark accuracy in classifying AD using neural models. 
We illustrate with two visualization techniques how these models capture unique linguistic features present in AD patients. 
We also discussed gender analysis.
Potential future work includes using more conversational context and implementing multi-class classification to differentiate among stages of AD. 
We also plan to apply this generalizable model to other similar neurological diseases, 
such as Diffuse Lewy Body disease and Huntington's disease~\cite{heindel1989neuropsychological}.

%% file: tex-files/supplementary.tex
\appendix
\section{Supplementary Materials}
\label{sec:supplemental}
\subsection{Training Details}
All models had a vocabulary size of $2396$, 
and used an Adam optimizer~\cite{kingma2014adam} with a learning rate of $1e^{-4}$. 
All gradient norms were clipped to $2.0$~\cite{pascanu2013difficulty,graves2013generating}.
For each model, the hyperparameters were tuned using the development set. 

\paragraph{CNN}
We used a $2$-D CNN.
Filter sizes of [$3$, $4$, $5$] were used with $128$ filters per filter size. 
Batch size was set to $128$, and a dropout~\cite{srivastava2014dropout} of $0.80$ was applied.

\paragraph{LSTM}
Our LSTM had $2$ layers with $128$ hidden units. 
Batch size was $32$, and a dropout of $0.70$ was used. 

\paragraph{CNN-LSTM}
Our CNN-LSTM model consisted of an LSTM on top of a CNN. 
The CNN had 100 filters per filter size of [$3$, $4$, $5$, $6$]. 
Embedding dimensions of $300$ were used. 
An LSTM with $300$ hidden units was used. 
Both dropout on the CNN and recurrent dropout~\cite{gal2016theoretically} on the LSTM used a dropout rate of $0.65$. 

\subsection{Dataset Details}
Each transcript in DementiaBank comes with automatic morphosyntactic analysis, 
such as standard part-of-speech tagging, description of tense, and markers for repetitions. 
This automatic tagging is identical to that done on other datasets, such as the CHILDES TalkBank, 
and is thus not specific to DementiaBank. 
This dataset features transcripts of $104$ different control patients, 
and $208$ different diagnosed dementia patients. 
There is a total of $1017$ Alzheimer's transcripts and $243$ control transcripts.
Each of these transcripts were then broken down by sentences and interruptions by the interviewer. 
We used each utterance by the patient as a data sample. 
Within the $14362$ utterance samples, $11458$ come from transcripts of Alzheimer's-diagnosed interviewees and $2904$ from those of control patients. 
Therefore, a majority-baseline classifier that always guesses \textit{AD-positive} will achieve an accuracy of $79.8\%$ in our dataset. 
Each utterance has a POS-tagged counterpart in the dataset.
A $80$/$10$/$10$ train/dev/test split was used for each setting.